\documentclass{llncs}
\usepackage{booktabs}
\usepackage{url}
\usepackage{hyperref}
\usepackage[table,xcdraw]{xcolor}
\usepackage{graphicx}
\usepackage{amssymb}
\usepackage{pifont}
\usepackage{color}
\usepackage{algorithm}
\usepackage{algorithmicx,algpseudocode}
\usepackage[normalem]{ulem}
\usepackage{textcomp,footnote}
\usepackage{amsmath,amssymb}
\usepackage[symbol]{footmisc}
\usepackage{subfigure}
\usepackage{caption}
\usepackage{xcolor}
\usepackage{color}
\usepackage{todonotes}

\begin{document}
	\algnewcommand\algorithmicforeach{\textbf{for each}}
	\algdef{S}[FOR]{ForEach}[1]{\algorithmicforeach\ #1\ \algorithmicdo}
	
	\title{Building robust prediction models for defective sensor data using Artificial Neural Networks}
	
	\author{Arvind Kumar Shekar\inst{1} \and Cl{\'a}udio Rebelo de S{\'a}\inst{2,3} \and Hugo Ferreira\inst{2} \and Carlos~Soares\inst{4}}
	
	
	
	\institute{
		Robert Bosch GmbH, Stuttgart, Germany\\
		\email{arvindkumar.shekar@de.bosch.com} \and
		LIACS, Leiden University, Netherlands\\
		\email{c.f.de.sa@liacs.leidenuniv.nl}
		\and
		INESC TEC, Porto, Portugal \\
		\email{hmf@inesctec.pt}
		\and
		Faculdade de Engenharia, Universidade do Porto, Portugal \\
		\email{csoares@fe.up.pt}
	}
	
	\maketitle              
	
	\begin{abstract}
		Predicting the health of components in complex dynamic systems such as an automobile poses numerous challenges. 
		The primary aim of such predictive systems is to use the high-dimensional data acquired from different sensors and predict the state-of-health of a particular component, e.g., brake pad. 
		The classical approach involves selecting a smaller set of relevant sensor signals using feature selection and using them to train a machine learning algorithm. 
		However, this fails to address two prominent problems: 
		(1) sensors are susceptible to failure when exposed to extreme conditions over a long periods of time; 
		(2) sensors are electrical devices that can be affected by noise or electrical interference. 
		Using the failed and noisy sensor signals as inputs largely reduce the prediction accuracy. 
		To tackle this problem, it is advantageous to use the information from all sensor signals, so that the failure of one sensor can be compensated by another. 
		In this work, we propose an Artificial Neural Network (ANN) based framework to exploit the information from a large number of signals. 
		Secondly, our framework introduces a data augmentation approach to perform accurate predictions in spite of noisy signals. 
		The plausibility of our framework is validated on real life industrial application from Robert Bosch GmbH. 
		
		
	\end{abstract}
	
	\section{Introduction}
	\label{sec:Introduction}
	
	Predicting the wear out of components is pivotal in various domains such as the automotive, health and aerospace industries~\cite{allred1998prognostic,ReussSAHH18,DBLP:conf/pkdd/ShekarBSSM17}.
	Robust and accurate predictions have a great potential for preventing unanticipated equipment failures and increasing productivity.
	With the recent widespread adoption of the Internet-of-Things (IoT), many sensor signals are now readily accessible for predicting the wear out of components.
	
	At Bosch, we often encounter datasets with several hundreds of sensor measurements and other calculated values from vehicles~\cite{DBLP:conf/pkdd/ShekarBSSM17}. These are used for predicting the health-state of a component. For example, in automotive applications, we can predict the wear out of an engine-coolant system using signals from different sensors such as torque, pressure, temperature and speed. 
	Traditional approaches select a small and predictive subset of these measurements (or attributes) by evaluating their relevance to the target (health-state) prediction~\cite{ChandrashekarS14,MolinaBN02}. 
	Several off-the-shelf algorithms, viz., Decision Trees \cite{quinlan2014c4}, Random forests \cite{breiman2001random}, Gaussian processes \cite{lazaro2010sparse} and Support Vector Machines (SVM's) \cite{libsvm}, were used on our fuel system data from different vehicles. Overall, we observed that all aforementioned algorithms selected a similar subset of attributes as the most relevant ones.
	
	A problem arises in the case when one or more of these selected (relevant) attributes are invalid due to malfunctioning sensors. 
	During malfunctioning, the sensors measurements are stuck at a constant value, e.g., zero, such cases are denoted as stuck-at-zero condition of the sensor \cite{elleithy2012innovations}. If such a malfunctioning sensor represents a relevant attribute for the target prediction,
	it leads to unreliable predictions. 
	It is therefore essential to train a model that does not rely on a fixed subset of attributes. 
	Additionally, sensors are electrical devices that are prone to be affected by noise. 
	For example, the magnetic field generated by the ignition system of a vehicle can affect other sensors~\cite{dziubinski2016electromagnetic}. Noisy sensors generate a few distorted measurements amidst valid values. 
	Using these distorted sensors readings can lead to erroneous predictions and raise false alarms by the wear out prediction model. 
	Industries spend millions of dollars to remove the noise from these signals \cite{redman1997data}. 
	However, manual data cleansing process is laborious, time consuming and prone to errors \cite{zhu2004class}.
	
	The first challenge is to generate a prediction model that is robust to missing attributes, i.e., stuck-at-zero condition.
	The second challenge is to ensure that the prediction model is robust against noisy attributes. 
	Solving these two problems are one of the foremost challenges that Bosch faces when predicting the health-state of the vehicle's components.
	For the aforementioned challenges, we propose:
	\begin{enumerate}
		\item A technique for building prediction models that are robust to faulty or missing attributes.
		\item A strategy for handling noise in the input attributes, that is built upon the data augmentation technique.
	\end{enumerate}
	
	To enhance the robustness of the predictions in spite of faulty attributes, we propose using prediction models that do not rely on a small set of signals. 
	Our approach is founded upon the Dropout technique, a well-known regularization technique used in the training of Artificial Neural Networks (ANN).
	Dropout randomly removes a few attributes during training.
	This forces the ANN to use more attributes during the training phase instead of relying on a single small subset of attributes. 
	Moreover, random dropping of the ANN units during training of the network simulates the situation of sensor failure in the real world. 
	To address the second challenge of noisy inputs, ANNs were trained with a certain magnitude of synthetically generated noise in the training data. 
	By replacing the values of the attributes in the training data with random values from a Gaussian distribution, we indirectly simulate the noisy behavior of the sensors. 
	This allows the ANN to learn the contributions of each feature for the output prediction amidst distorted inputs. 
	Bosch provided a labeled dataset related to the health-state of the fuel system. 
	Using this automotive data, we tested the robustness of our framework on a real world scenario. 
	
	\section{Related Work}
	As elaborated in the previous section, first we aim to perform predictions based on a large subset of attributes to avoid incorrect predictions during sensor failure. 
	Secondly, we aim to augment the training data to enhance the ability of the network to be able to identify relevant patterns amidst noisy input data. 
	
	\textit{Preprocessing} techniques for handling noisy and missing input attributes have been of great interest in the data mining community \cite{redman1997data,zhu2004class,wang1995framework,maletic2000data}. The aforementioned methods have their own strengths and weaknesses. However, in real world applications, we do not know the type of noise that can interfere with the sensor measurements. 
	As mentioned in Section \ref{sec:Introduction}, valid sensor measurements can be stuck-at-zero \cite{elleithy2012innovations} in case of malfunction. Applying imputation techniques to extrapolate these values as in the case of a missing value problem is not desirable. 
	Hence, it is not a pragmatic solution apply these data preprocessing techniques in real world applications \cite{zhu2004class}. 
	
	\textit{Feature Selection} algorithms predominantly focus on selecting a set of attributes relevant for the prediction task \cite{MolinaBN02,ChandrashekarS14,DBLP:conf/pkdd/ShekarBSSM17}. 
	The recent work of \textit{Relevance and Redundancy} ranking \cite{DBLP:conf/pkdd/ShekarBSSM17} is a feature ranking framework that has experimentally shown to be robust amidst noisy target labels. 
	However, we focus on building prediction model using a large number of attributes to enhance robustness of predictions. Secondly, our application scenario involves noisy input attributes and not noisy target labels. 
	
	\textit{Multi-view learning} algorithms perform predictions based on multiple attribute subsets. In the case of a failed attribute in one subset, the predictions can be supported by attributes from other subsets. 
	However, existing multi-view approaches \cite{DBLP:conf/dawak/ShekarSM17,oza1999dimensionality} do not discuss the effect of faulty input attributes. Nor are they as resistant to multiple sensor failures that can occur over all of the attribute subsets.  
	
	\textit{Pruning of Decision trees} was introduced to avoid over fitting to noisy training data \cite{quinlan2014c4}. 
	As classifiers learned from noisy data have less accuracy, pruning may have very limited effect in enhancing the system's performance, especially in the situation that the noise level is relatively high \cite{zhu2004class}.
	
	\textit{Dropout} technique in ANNs is similar to the idea of pruning in decision trees. The regularization technique of dropout aims to eliminate random units of the neural network to avoid over fitting. However, in this work we use this regularization technique because performing dropout in the inputs is analogous to the real world scenario of sensor failure. 
	
	The technique of \textit{adding noise} to the training data is reported to enhance the generalization of ANNs by forcing more hidden units to be used \cite{sietsma1991creating}. 
	Hence, to address the second problem of noisy input attributes, we use artificially generated noise in the training data. 
	By training the prediction model with artificially injected noise in the training data, we aim to enhance the prediction model's ability to identify relevant patterns amidst noise in the real world scenario. 
	
	Hence, in contrast to the preprocessing techniques, our work aims to challenge the prediction model during training phase by forcing it to learn relevant patterns amidst noise. 
	
	\section{Problem Definition} \label{sec:problem}
	As explained in Section \ref{sec:Introduction} we address the first problem building prediction models with inputs obtained from malfunctioned sensors. Hence, we begin with the formal definition of a faulty sensor. 
	\begin{definition}{Malfunction of sensors} \label{def:malfuntion}\newline
		Assume a $d$-dimensional attribute space $\mathcal{F}= \{a_1,\cdots,a_{d}\}$, where a subset of sensors $M\subset\mathcal{F}$ are defective. 
		This means that each attribute $a\in M$ is stuck at zero and continuously generates null values. 
	\end{definition}
	The second problem being noise in the sensor data, we formally define the behavior of a noisy sensor. 
	\begin{definition}{Noisy sensor}\label{def:noisy}\newline
		Assume a subset of sensors $N\subset\mathcal{F}$, that are subjected to intermittent deviations or disturbances. 
		This means that the random instances of attribute $a\in N$ fluctuates to absurd values and deviates from the actual measurements. 
	\end{definition}
	
	We denote the accuracy of a prediction model trained using the attribute space as $acc:\mathcal{F}\mapsto\mathbb{R}$. 
	We focus on enhancing the robustness of the predictions such that, in the event of a sensor failure, we aim to obtain an accuracy greater than or equal to that of a prediction model with all valid measurements.
	$$acc(\mathcal{F}\mid|M|<1) \leq acc(\mathcal{F}\mid|M|>1)$$
	Similarly, in the case of a noisy sensor,
		$$acc(\mathcal{F}\mid|N|<1) \leq acc(\mathcal{F}\mid|N|>1).$$
	\section{Artificial Neural Networks}
\begin{figure}
		\vspace{-3mm}
		\center
		\includegraphics[width=0.7\textwidth]{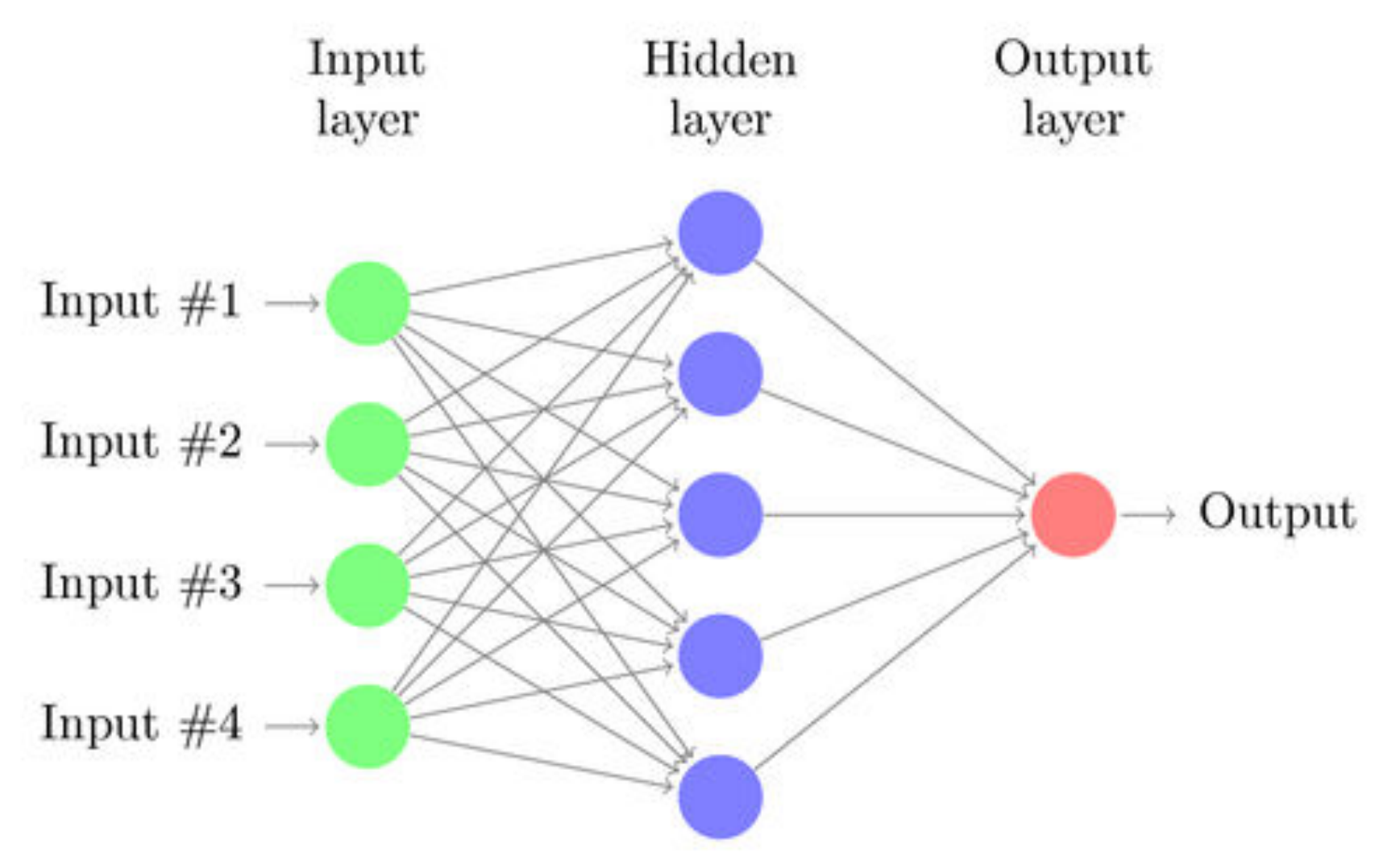}
		\caption{Schema of an artificial neural network. Image Source \cite{ANNimage} }
		\label{fig:ANN}
		\vspace{-4mm}
	\end{figure}

	To obtain a deeper understanding about the dropout technique, it is necessary to revisit the basics of ANNs. 
	ANNs are machine learning algorithms inspired by the biological nervous system and are capable of identifying complex non-linear relationships. 
	Information is processed using a set of highly interconnected nodes, also referred to as neurons. 
	A network of weighted nodes are stacked into multiple layers. 
	At each node, an activation function combines the weights  
	into a single value. 
	This can effectively limit the signal propagation to the next layers. 
	These weights, therefore, enforce or inhibit the activation of the network’s nodes. This process is comparable to feature selection. 
	Additionally, ANN's require minimal attribute engineering for classification \cite{DBLP:journals/neco/Baxt90,DBLP:journals/cacm/WidrowRL94} and regression~\cite{DBLP:journals/nn/RefenesZF94} problems. This enables ANNs to autonomously identify distinct patterns in the input attributes amidst noise. 
	Hence, with embedded feature selection and the ability to identify distinct patterns with minimal preprocessing, we chose ANNs as an ideal candidate for our experiments. 
	The ANN architecture is typically split into three types of layers: one input layer; one or more hidden layers; and one output layer (c.f. Figure \ref{fig:ANN}).
	The input layer consumes the data.
	This layer connects to the first hidden layer, which in turn connects either to the next hidden layer (and so on) or to the output layer.
	The output layer returns the ANN’s predictions.
	
	There are two main types of ANNs based on the flow of information, referred as Feed-forward Neural Network (FNN) and Recurrent Neural Network (RNN)~\cite{DBLP:conf/emnlp/ChoMGBBSB14}.
	In FNNs, the flow of information through the hidden layers is acyclic. 
	On the other hand, with RNN, the flow of information in the hidden layers can be bi-directional or cyclic. 
	FNNs have been used in many different domains such as the prediction of medical outcomes \cite{TU19961225}, environmental problems \cite{Maier2000101}, stock market index predictions \cite{Moghaddam201689} and the wear out of machines \cite{BenAli2015150}. 
	Considering its wide usage in applications analogous to ours, in this work we choose to use FNNs for building our prediction framework. 
	Using the FNN, we aim to address the first challenge defined in Section \ref{sec:Introduction}. 
	That is, to build prediction models that are robust to faulty attributes (c.f. Definition \ref{def:malfuntion}). 
	For this we apply the concepts of dropout, which we describe next. 
	
	\subsection{Dropout} \label{subsec:dropout}
\begin{figure}
		\vspace{-3mm}
		\center
		\includegraphics[width=0.7\textwidth]{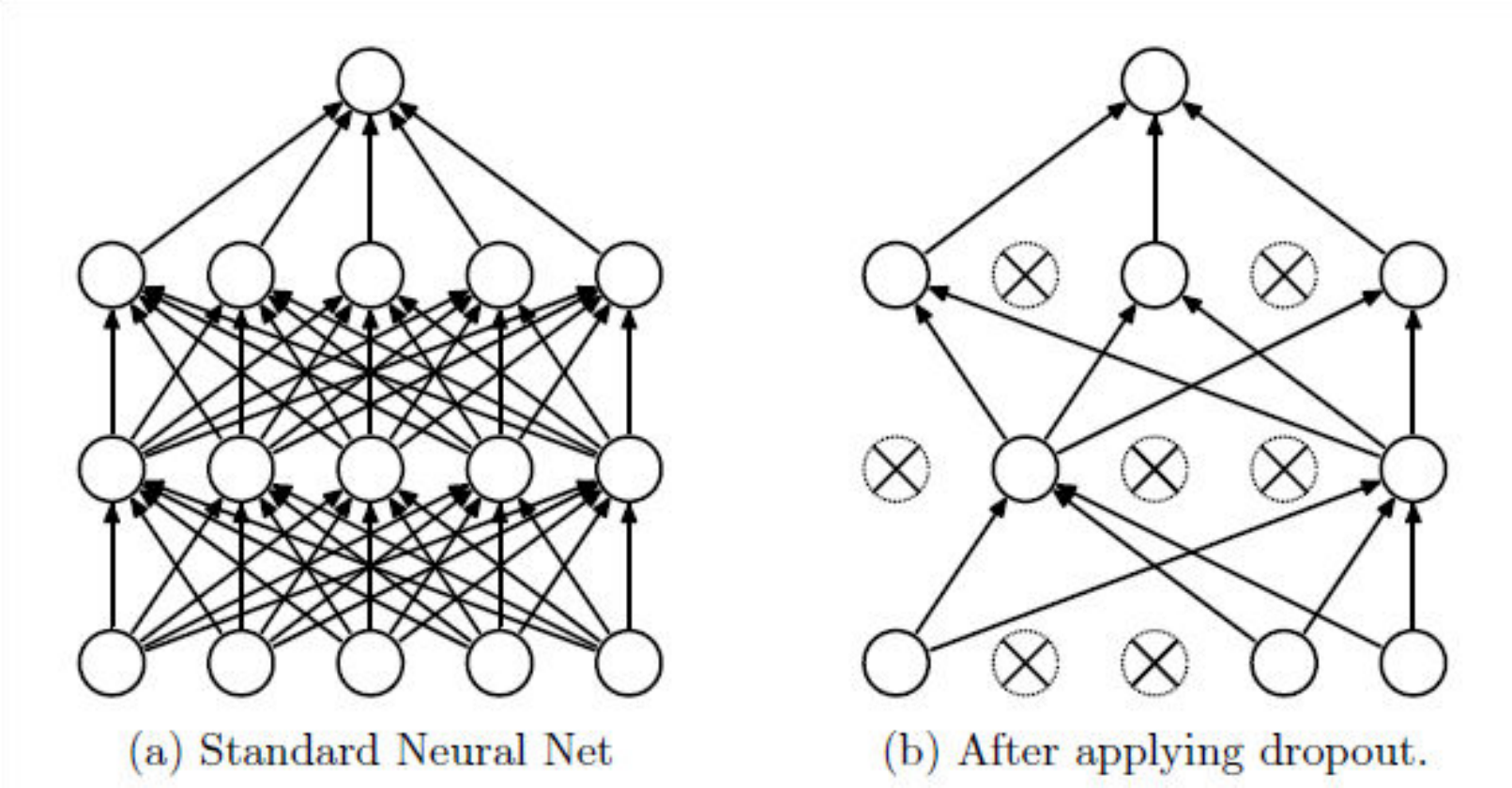}
		\caption{Example of Dropout used in ANN (Image Source: \cite{DBLP:journals/jmlr/SrivastavaHKSS14})}
		\label{fig:dropout}
		\vspace{-3mm}
	\end{figure}

	Dropout is proven to be an effective regularization technique for ANNs~\cite{DBLP:journals/jmlr/SrivastavaHKSS14}. Technically, it prevents the units from co-adapting too much and consequently avoids over-fitting while training the network.
	Dropping or removing an unit implies that both the input and output connections of the neuron are disconnected. 
	In Figure \ref{fig:dropout}, we provide an illustration of networks with fully connected and dropped out units.  
	The principal idea of dropout involves removing random units from a layer (both hidden and visible) by setting its activation function to zero. 
	That is, when applied on the input layer, the activations of selected neurons are nullified. Therefore, application of dropout on input layer is analogous to the sensor failure in real world scenario (c.f. Definition \ref{def:malfuntion}). 
	By training the ANNs with dropout, we indirectly aim to make the network aware of these failures. 
	
	The abstract concept of Dropout \cite{breiman2001random} sounds very similar to the ensemble technique used by Random forests. 
	Random forest aggregate prediction results from the multiple views of the data based on a number of decision trees that use randomly selected subsets of attributes.  
	Similarly, Dropout networks essentially train different networks on multiple subset of the attributes. 
	However, on a closer look into the details, there are considerable differences between both (c.f. Table \ref{table:differenceRfVsDrop}). 
	\begin{table}
		\vspace{-12mm}
		\centering
		\caption{Differences between Random forest ensembles and Dropout Networks \cite{warde2013empirical,jaquescomparison}}
		\label{table:differenceRfVsDrop}
		\resizebox{\columnwidth}{!}{ 
			\begin{tabular}{c|c}
				\toprule
				Random Forest                                                                                                                                  & Dropout Network                                                                                                                                              \\ \hline
				\begin{tabular}[c]{@{}c@{}}A large number of  decision trees are trained using\\ randomly selected attribute subsets in parallel.\end{tabular} & \begin{tabular}[c]{@{}c@{}}It is an inherently serial process, where\\neurons are dropped out as each training\\sample is processed.\end{tabular} \\ \hline
				All data samples  are used.                                                                                                                    & A single sample is used to train a model.                                                                                                                    \\ \hline
				Each tree has independent parameters.                                                                                                          & \begin{tabular}[c]{@{}c@{}}The parameters are shared between networks with\\different neurons dropped.\end{tabular}                                          \\ \hline
				Arithmetic mean to combine the results.                                                                                                        & Equally weighted geometric mean to combine results.                                                                                                          \\ \bottomrule
			\end{tabular}
		}
		\vspace{-8mm}
	\end{table}
	
	Dropping random neurons in each iteration enables every hidden unit to learn to identify relevant patterns from a randomly chosen sample of neurons of the preceding layer. 
	This makes each hidden layer robust and drives them to create useful features on their own without requiring that the next layers correct their mistakes~\cite{DBLP:journals/jmlr/SrivastavaHKSS14}.
	Recent study also shows that Dropout networks are comparatively more accurate than Random forest for multi-class classification problems \cite{jaquescomparison}. 
	
	\subsection{Data Augmentation}\label{subsec:dataAugmentataion}
	As explained in Section \ref{sec:Introduction}, in automotive applications, exposing the sensors to harsh-environmental conditions over a prolonged period of time can cause the sensor values to be distorted due to electrical or magnetic interference \cite{dziubinski2016electromagnetic}. 
	Hence, training the machine learning models to identify relevant patterns irrespective of noisy attributes is of paramount importance. 
	To mimic the problem of noisy sensors (c.f. Definition \ref{def:noisy}) in real world applications, we performed data augmentation on our training data.
	Data augmentation is a concept introduced from the literature of image classification~\cite{arandjelovic2012three}. 
	It involves transforming the original data (e.g., rotation, zoom, rescaling and cropping) to avoid over-fitting~\cite{2017arXiv171204621P}. For example, to build text-to-speech models, the data is collected from unfiltered Web pages with errors. 
	Rather than using the large unstructured data for learning useful patterns, a small corpus of structured data is extracted and augmented. 
	It is then used to train the machine learning model. This technique has also proven to be effective on unfiltered data that contain errors \cite{2017arXiv171204621P}.
	
	We adopt the concept of data augmentation and tailor it to address our second challenge (c.f. Section \ref{sec:Introduction}), i.e., noisy attributes. We replace random attributes in the dataset with noise. 
	That is, we deliberately introduce noise to the original training data and then train our models using this transformed dataset. 
	In practical terms, the values of a randomly selected subset of attributes in each instance is replaced with random values obtained from a Gaussian distribution with mean zero and standard deviation of one, i.e., $\mathcal{N}(0,1)$. 
	Hence, by training the models with certain levels of noise, we enhance their robustness against sensor failures in the real world.
	
	\section{Methodology} \label{sec:methodology}
	In Section \ref{subsec:dropout} and \ref{subsec:dataAugmentataion} we justified the use of dropout and data augmentation to address the problems we are confronted with (c.f Section \ref{sec:problem}). 
	The theoretical concept of dropout and data augmentation emulates the real life situation of sensor failure and noise respectively. 
	However, its practical application raises two major questions, 
	\begin{enumerate}
		\item What is the magnitude of dropout to be used?
		\item What is the level of augmentation to be applied for the transformation of the training data?
	\end{enumerate}
	For this, we train multiple models with different levels of input dropout and data augmentation. 
	These models are tested upon test data and we observe the prediction accuracy on it as a quality measure.  
	We explain the finer details based on the dataset we use.
	\subsection{Dataset}
	In this work, we apply the proposed methodology to an automotive dataset. 
	We are provided with a high-dimensional attribute space $\mathcal{F}= \{a_1,\cdots,a_{149}\}$ of 149 attributes and 4 million instances. 
	The attributes are obtained from various sensor sources present in the vehicles.
	It also include signals that are calculated in the vehicle hardware using the sensor measurements. 
	The goal is to predict the target classes that represent the health-state of an automotive fuel system. 
	Therefore, we are provided with the target labels ($Y$) of nominal values and the dataset\footnote[3]{Code and data: \url{https://figshare.com/s/d5bcd9b4269afa642e53}} is denoted as $\mathcal{D}=\{\mathcal{F}, Y\}$. 
	
	Table \ref{table:classDistribution} shows the distribution of the different classes in the dataset. 
	As the data for each health state was obtained from different vehicles, each instance can be seen as a snapshot of the fuel system.
	In other words, the dataset is not a time-series and health-states are therefore not correlated in time.
	In such stationary datasets, FNN's are a preferable choice in comparison to RNN's. 
	\begin{table}
		\vspace{-6mm}
		\caption{Distribution of the classes in the dataset}\label{table:classDistribution}
		\centering
		\resizebox{0.5\textwidth}{!}{%
			\begin{tabular}{lrr}
				\toprule
				Class & Health state& Class distribution \\
				\midrule 
				Class 1 & 0\% 	& 9.96\% \\
				Class 2 & 10\% 	& 13.98\% \\
				Class 3 & 20\% 	& 3.6\% \\
				Class 4 & 40\% 	& 4.6\% \\
				Class 5 & 60\% 	& 12.8\% \\
				Class 6 & 80\% & 47.06\%\\
				Class 7 & 100\% & 7.9\%\\ \bottomrule
				\hline
			\end{tabular}
		}
		
	\end{table}
	
	The dataset is split into two parts for training and testing purposes based on the chronology of the data collection. 
	That is, training is performed using the data collected on a specific time of the year (e.g., January) and the testing is performed on a dataset collected from a different time (e.g., August). 
    Both train and test datasets were standardized by subtracting the mean and dividing by the standard deviation. This is also referred as z-score or a standard score. 
	
	The training dataset is used to train 7 different networks, each with different magnitude of input dropout. 
	For example, $Model~D2$ denotes an ANN model with a dropout of 20 nodes in the input layer.
	Similarly, we instantiate multiple networks ($Model~D2,Model~D4,\cdots,D14$) with varying dropout levels of $20,40\ldots,140$ attributes respectively. 
    
	Given an ANN architecture and a dropout level, the dropout can be applied between any two consecutive layers.
	Nevertheless, we aim study the influence of dropout between the input and the first hidden layer. 
	This implicitly means that each model is trained to predict with a different number of faulty sensors.
	However, a constant dropout rate of $50\%$ was still used in the hidden layers for regularization purposes. 
	To drop one neuron, is technically setting the activations of this neuron to zero. 
	Hence, we transform the original dataset to mimic the dropout process in the input layer by setting its value to zero. 
	The reason for setting attribute values to zero instead of using the dropout in the input layer of the ANNs is that it allows us to simulate an equivalent dropout in the test dataset as well. 
	The corresponding test datasets are denoted as $DTest2,DTest4,\ldots,DTest14$. Moreover, this experimental setting is comparable to the problem of failed sensor that is stuck-at-zero (c.f. Definition \ref{def:malfuntion}). 
	For simplicity we refer to the original train and test dataset as $D0$ and $DTest0$ respectively.
	The goal of the experiment is to identify the level of dropout that has the maximal accuracy on the unseen test data. 
	\vspace{-1mm}
	\begin{algorithm}
		
		\renewcommand{\algorithmicrequire}{\textbf{Input:}}
		\renewcommand{\algorithmicensure}{\textbf{Output:}}
		\caption{Algorithm for injection of noise into data}
		\begin{algorithmic}[1] 
			\Require  $\mathcal{F}, \alpha \in \{0,20,40,\cdots,140\}$
			\State $\mathcal{I}=\{1,...,149\}$ \Comment{Set of attribute indices}
			\ForEach {$Instance~i$}
			\State $Select~random~subset~of~attribute~indices~\mathcal{I}'\subset \mathcal{I},~where,~\mid\mathcal{I}'\mid=\alpha$
			\State $Replace~instance~i~of~attribute~a_j\in\mathcal{F}\mid\forall j \in \mathcal{I}'~with~values~from~\mathcal{N}(0,1)$ \label{line:Replace}
			\EndFor
		\end{algorithmic}	
		\label{alg:noiseApproach}
	\end{algorithm}
	\vspace{-3mm}
	
	In the case of augmentation, injecting noise in all instances of a single subset of attributes is not challenging for the network because the ANN will simply neglect these attributes during training by inhibiting the corresponding network nodes. 
	Hence, for each instance of the attribute space $\mathcal{F}$, a random attribute subset of size $\alpha \in \mathbb{Z}$ (where, $0\leq\alpha<|\mathcal{F}|$) is selected and replaced with random values from a Gaussian Distribution (c.f. Algorithm \ref{alg:noiseApproach}). 
	In our experiments, $V0,V2,V4...,V14$ denote different variants of training data with $\alpha \in \{0,20,40\cdots,140\}$ respectively. 
	For example, $V2$ represents a dataset where 20 random attributes of the training data are replaced by random numbers from a Gaussian distribution for each instance. 
	By applying the transformation, our goal is to imitate the real world scenario of noisy sensors and analyze the influence of different noise levels in the input attributes. The corresponding transformation is also applied to the test data and is denoted as $VTest0,VTest2,VTest4,\cdots,VTest14$.  

	In electrical applications, white noise is also a commonly observed anomaly in the sensor measurements. Hence, we also generate test datasets with white noise, i.e., $WTest2,WTest4,...,WTest14$. For the generation of data with white noise, we follow the same sequence of steps explained in Algorithm \ref{alg:noiseApproach}. 
	However, instead of replacing (c.f. Line \ref{line:Replace} in Algorithm \ref{alg:noiseApproach}), we add valid measurements in an instance with random values from $\mathcal{N}(0,1)$. 
	As a rule of thumb, all experiments in the forthcoming section will use a FNN architecture with: an input layer of 149 neurons, three hidden layers of 128, 256 and 128 neurons, and an output layer of 7 neurons. 
	
	\section{Experimental Results}
	As described in Section~\ref{sec:methodology}, we have 4 types of data: train data with dropout, test data with dropout, train data with noise and test data with noise.
	To test the influence of dropout and noisy attributes on the test data accuracy, we begin with individual analysis of each technique. 
	
	\subsection{Input drop}
	\label{subsec:inputdropExp}
\begin{figure}
		\vspace{-6mm}
		\centering
		\includegraphics[trim=0 0 0 0.58cm, clip, scale=0.7]{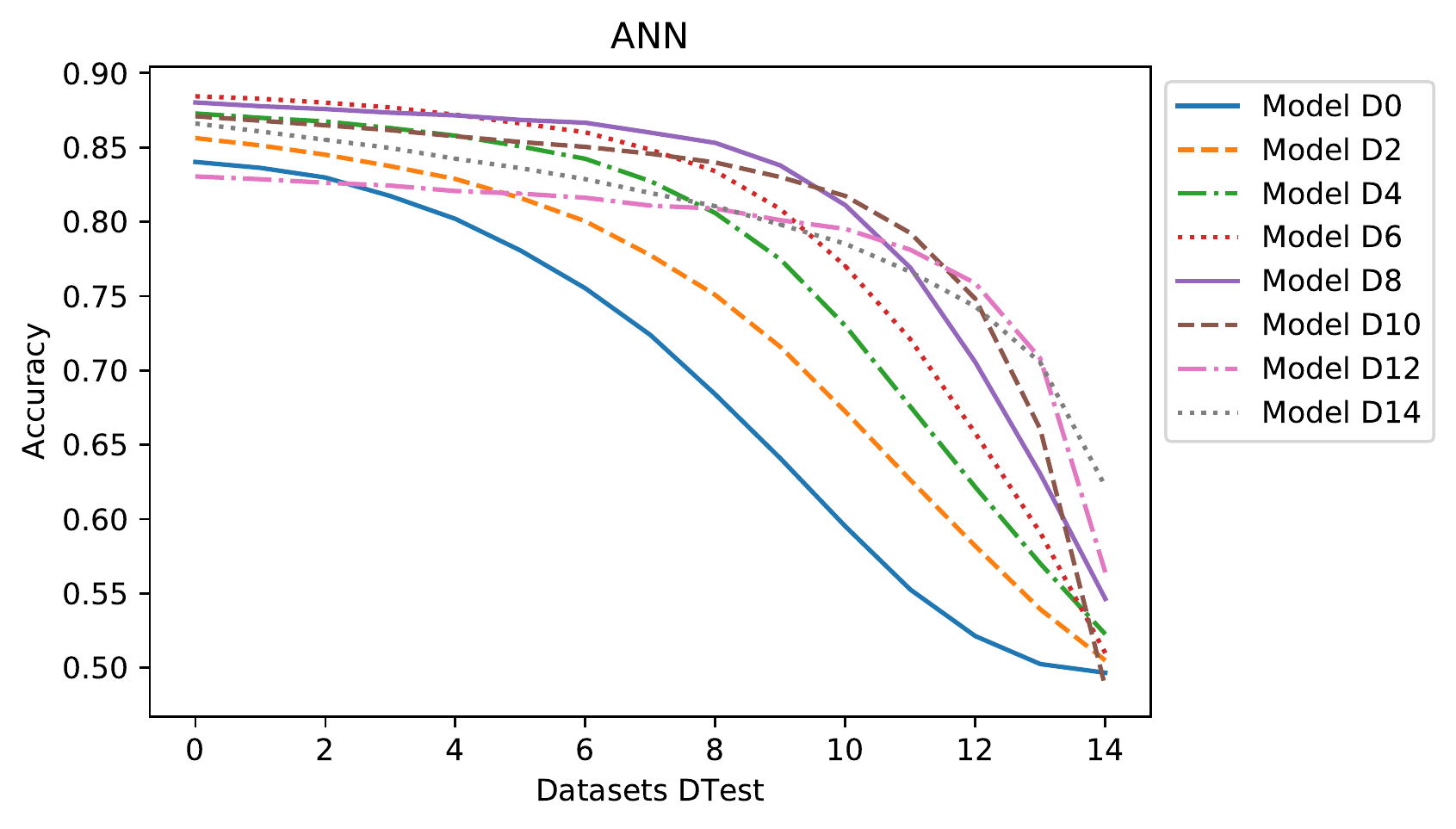}
		\vspace{-2mm}
		\caption{Accuracy (y axis) of different models trained using input drop data. The accuracy was calculated for each test data (x axis) with different levels of dropout ($DTest0,...,DTest14$).}
		\label{img:inputdrop_inputdrop}
		\vspace{-6mm}
	\end{figure}

	In this section, we experiment using ANN networks trained with different levels of dropout.
	In the first experiment, we trained multiple networks with the datasets $D0, D2,D4,\ldots,D14$. 
	Each of these models were then evaluated on all test datasets that were subjected to the same input drop process which are denoted as $DTest0, DTest2,..., DTest{14}$ respectively. 
	The results are illustrated in Figure~\ref{img:inputdrop_inputdrop}.
	The network trained with the original data, i.e., $Model~D0$, is accurate when tested on datasets with low or no dropout, i.e., $DTest0$ and $DTest2$. 
	After this point onwards, its accuracy declines steeply with an increasing number of dropped inputs in the test dataset, until it reaches an accuracy of 0.5 for $DTest14$.
	Interestingly, we observe that the models which were trained on datasets with a larger number of dropped inputs, are comparatively more robust to test data with a large number of dropped inputs. 
	Moreover, they also maintain a high accuracy on test datasets that have more dropped inputs than the one used for training. From the experimental analysis, we observe that the average of all test data accuracies using $Model~D8$ is higher in comparison to the other models. 
	It is therefore much more robust than $Model~D0$ with no dropped units. 
	Let us assume $Model~D8$ is used in a real world scenario to predict the health of the fuel system. In-spite of the failure of 100 sensors ($DTest10$) that are used as input attributes for the prediction model, the predictions will still have an approximate accuracy of 0.85. Hence, the idea of dropout helps us to tackle the problem of failed sensors in the real world prediction systems (c.f. Section \ref{sec:problem}). 
	
	\subsection{Input noise}
	\label{subsec:inputnoise}
\begin{figure}
		\vspace{-4mm}
		\centering
		\includegraphics[trim=0 0 0 0.58cm, clip, scale=0.7]{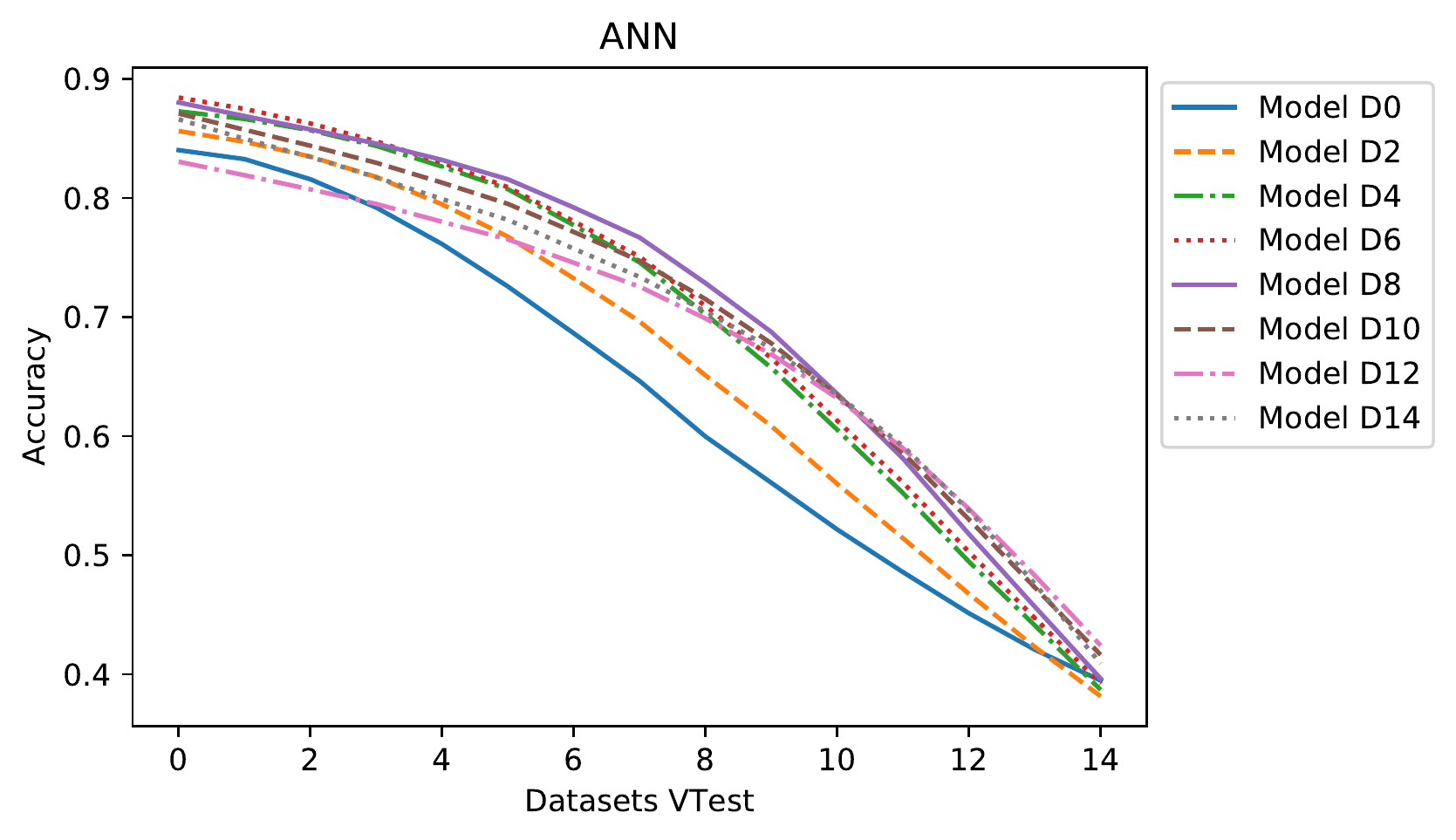}
		\vspace{-2mm}
		\caption{Accuracy (y axis) of different models trained on input drop data. The accuracy was measured for each test data (x axis) with different levels of noise ($VTest0,...,VTest14$).}
		\label{img:inputdrop_noisedrop}
		\vspace{-0.5cm}
	\end{figure}

	The above dropout experiment does not solve our problem completely because, a noisy sensor will not be seen as missing data. Instead, it will give us a wrong measurement.
	For this reason, we did a second experiment where we test the input dropout models, i.e., $Model~D0,Model~D2,...,D14$, on scenarios where the data has faulty measurements.
	That is, we tested the dropout models on test data obtained from the input noise approach, viz., $VTest0, VTest2, \ldots, VTest14$.
	The behavior of the models are visually represented in Figure~\ref{img:inputdrop_noisedrop}.
	In comparison to the previous experiment (c.f. Figure \ref{img:inputdrop_inputdrop}), all the models have worser performances because the decline in accuracy happens much earlier in Figure~\ref{img:inputdrop_noisedrop}. This is not surprising because the training was performed with dropout technique without noise and the testing was performed with noisy data. Hence, the network is unaware of the noise in the test data. 
	Nevertheless, by comparing the behavior of $Model~D0$ with $Model~D8$ and $D10$ we observe that training models with input drop is helping them to be more robust to noisy measurements and $Model~D8$ was having the best performance in terms of accuracy. 

\begin{figure}
		\vspace{-6mm}
		\centering
		\includegraphics[trim=0 0 0 0.58cm, clip, scale=0.7]{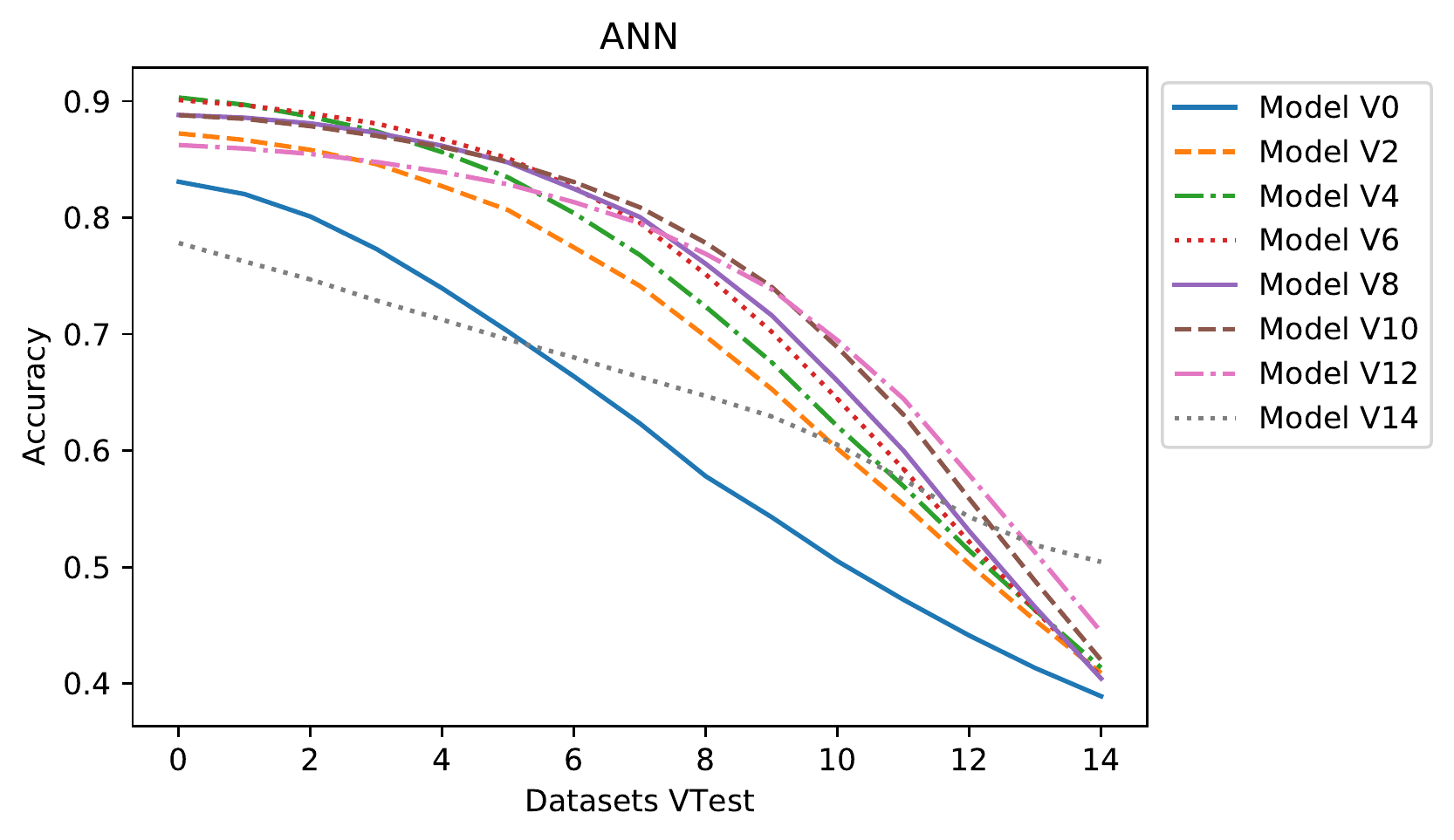}
		\vspace{-2mm}
		\caption{Accuracy (y axis) of different models trained on input noise data. The accuracy was measured for each test data (x axis) transformed with the same input noise approach, with different levels of noise ($VTest0,...,VTest14$).}
		\label{img:noisedrop_noisedrop}
		\vspace{-4mm}
	\end{figure}

	To make the network aware of noisy attributes, we perform a third experiment. 
	In the third experiment, we trained our models with the augmented dataset variants that include different levels of noise in the input data, i.e., $V0, V2,\ldots,V14$. The corresponding networks trained using these datasets are denoted as $Model~V0, Model~V2,...,Model~V14$.
	These models were validated on the test data $VTest0, VTest2,\cdots,VTest14$ that underwent a similar transformation (c.f. Algorithm \ref{alg:noiseApproach}).
	The results are plotted in Figure~\ref{img:noisedrop_noisedrop}.
	
	In Figure~\ref{img:noisedrop_noisedrop} we observe that $Model~V6$ and $V8$ have very similar behaviors. For example, $Model~V8$ is able to predict with an accuracy of 0.88 even when 40 sensors measurements are noisy.
	This represents around $25\%$ of the entire set of inputs.
	On the other hand, on test datasets with higher levels of noise, like $VTest14$, $Model~V6$ and $V8$ are unable to predict with high accuracy.
    
	Moreover, when comparing Figures~\ref{img:inputdrop_noisedrop} and \ref{img:noisedrop_noisedrop}, the results indicate that the best way to deal with noisy sensors is by training the ANN with reasonable levels of noise.
	This makes the models more robust to defective sensor data in real world.
	
    \begin{figure}[H]
		\vspace{-2mm}
		\centering
		\includegraphics[trim=0 0 0 0.58cm, clip, scale=0.7]{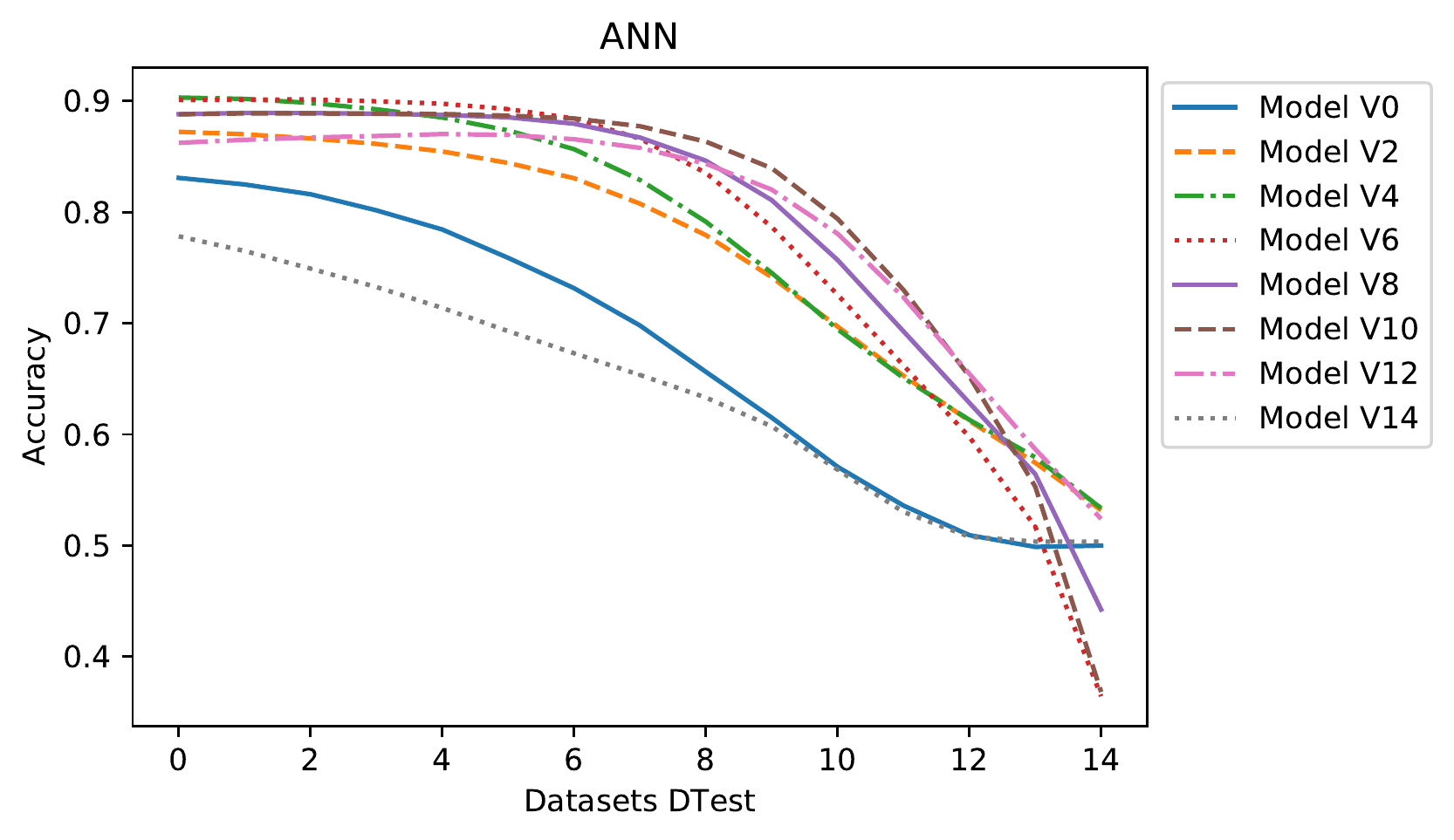}
		\vspace{-2mm}
		\caption{Accuracy (y axis) of networks trained using various levels of noise in training data and tested on datasets with varying levels of input dropout.}
		\label{img:ann_Dropnoise_Dropout} 
	\end{figure}
		Practically, our idea of injecting noise involves replacing the instances of the attribute space with random values from a Gaussian distribution. This also includes zeros. For this reason, the noise models trained on data $V2,...,V14$ also performs with a high accuracy on test datasets with input dropouts (c.f. Figure \ref{img:ann_Dropnoise_Dropout}). Also here, we observe that $Model~V8$ and $Model~V10$ have the best quality in comparison to the model trained with no random noise ($Model~V0$).

\begin{figure}
		\vspace{-2mm}
		\centering
		\includegraphics[trim=0 0 0 0.58cm, clip, scale=0.7]{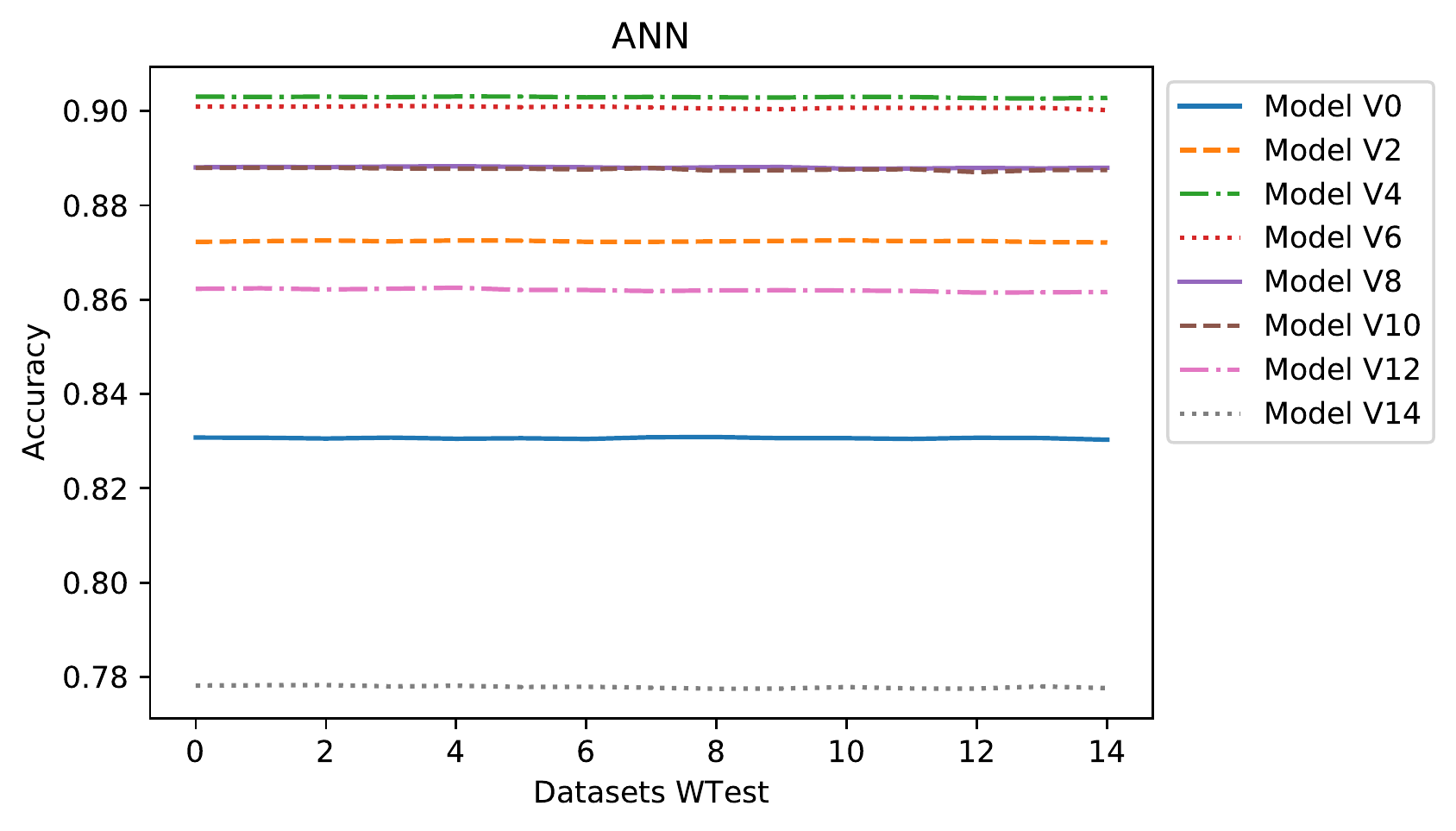}
	\vspace{-2mm}
		\caption{Accuracy (y axis) of networks trained using various levels of random noise in training data and tested on datasets with varying levels of white noise.}
	\label{img:ann_Dropnoise_DropWhiteNoise} 
	\end{figure}

	Similarly, these models were robust on test data with white noise. For example, in Figure \ref{img:ann_Dropnoise_DropWhiteNoise}, for test data with extreme levels of white noise, i.e., $WTest14$, the accuracy of the models trained with our random noise (e.g., $Model~V8$) is better in comparison to model trained using the original data ($Model~V0$). 
	
Overall, our observation is that our proposed idea of injecting random noise in the  instances of random features (c.f. Algorithm \ref{alg:noiseApproach})  enhance the robustness of the prediction model with malfunctioning and noisy sensors as inputs. 

\section{Conclusions and Future works}
	Bosch faces the challenge of generating prediction models with noisy and defective input attributes for applications such as predictive diagnostics.
	The models initially developed by Bosch using different classification algorithms produced very accurate results.
	However, a closer analysis showed that all these different prediction models relied on the same set of sensor data.
	Performing predictions with a single set of relevant sensor were not robust in the presence of faulty sensor data. 
	Hence, we proposed and tested two approaches to tackle this problem. 
	One approach (Input drop) uses the Dropout technique from ANNs in the input layer to make the model more robust against defective sensors.
	The second approach (Input noise) introduces noise into the training datasets, which can be seen as a way of simulating the noisy sensors.
    
	Based on our observations, the best level of dropout is between 60 to 80 attributes (i.e., between $40\%$ and $50\%$ of the attributes).
    As for the right level of augmentation, results indicate that model $V6$ (i.e., around $40\%$ of attributes) is ideal in terms of noisy and missing sensor data.
    
	While the major advantages of ANN are the effective and efficient modeling of complex non-linear systems, one downside is that, training a model usually incurs high computational and storage costs.
    On the other hand, once an ANN is trained, it requires little effort to process the data.
    This way, such a system could be implemented in the vehicles in a simple way. As future work, we intend to study if this approach can be generalized to other application domains, where sensor data are partially missing or faulty.
    

	\section*{Acknowledgments}
	This research has obtained funding from the Electronic Components and Systems for European Leadership (ECSEL) Joint Undertaking, the framework programme for research and innovation Horizon 2020 (2014-2020) under grant agreement number \emph{662189-MANTIS-2014-1}. We gratefully acknowledge the support of NVIDIA Corporation with the donation of the Titan X Pascal GPU used for this research.
	
	\bibliographystyle{abbrv}
	\bibliography{rul}
	
\end{document}